\documentclass[review]{elsarticle}
\usepackage{graphicx}
\usepackage{subfigure}    
\usepackage{amsmath}
\usepackage{amssymb}
\usepackage{multirow}
\usepackage[justification=centering]{caption}
\usepackage{bm}

\journal{Journal of \LaTeX\ Templates}

\begin{document}
\begin{sloppypar}
\begin{frontmatter}

\title{BB-GCN: A Bi-modal Bridged Graph Convolutional Network for Multi-label Chest X-Ray Recognition \tnoteref{mytitlenote}}

\author[mymainaddress,mysecondaryaddress]{Guoli Wang}
\author[mymainaddress,mysecondaryaddress]{Pingping Wang}
\author[mymainaddress,mysecondaryaddress]{Jinyu Cong}
\author[mymainaddress,mysecondaryaddress]{Kunmeng Liu}

\author[mymainaddress,mysecondaryaddress]{Benzheng Wei\corref{mycorrespondingauthor}}
\cortext[mycorrespondingauthor]{Corresponding author}
\ead{wbz99@sina.com}

\address[mymainaddress]{Center for Medical Artificial Intelligence, Shandong University of Traditional Chinese Medicine, Qingdao, 266112, China}
\address[mysecondaryaddress]{Qingdao Academy of Chinese Medical Sciences, Shandong University of Traditional Chinese Medicine, Qingdao, 266112, China}

\begin{abstract}
Multi-label chest X-ray (CXR) recognition involves simultaneously diagnosing and identifying multiple labels for different pathologies. Since pathological labels have rich information about their relationship to each other, modeling the co-occurrence dependencies between pathological labels is essential to improve recognition performance. However, previous methods rely on state variable coding and attention mechanisms-oriented to model local label information, and lack learning of global co-occurrence relationships between labels. Furthermore, these methods roughly integrate image features and label embedding, ignoring the alignment and compactness problems in cross-modal vector fusion.
To solve these problems, a Bi-modal Bridged Graph Convolutional Network (BB-GCN) model is proposed. This model mainly consists of a backbone module, a pathology Label Co-occurrence relationship Embedding (LCE) module, and a Transformer Bridge Graph (TBG) module. Specifically, the backbone module obtains image visual feature representation. The LCE module utilizes a graph to model the global co-occurrence relationship between multiple labels and employs graph convolutional networks for learning inference. The TBG module bridges the cross-modal vectors more compactly and efficiently through the GroupSum method.
We have evaluated the effectiveness of the proposed BB-GCN in two large-scale CXR datasets (ChestX-Ray14 and CheXpert). Our model achieved state-of-the-art performance: the mean AUC scores for the 14 pathologies were 0.835 and 0.813, respectively.
The proposed LCE and TBG modules can jointly effectively improve the recognition performance of BB-GCN. Our model also achieves satisfactory results in multi-label chest X-ray recognition and exhibits highly competitive generalization performance.
\end{abstract}


\end{frontmatter}


\section{Introduction}
\begin{figure*}[htbp]
\centering
\includegraphics[width=0.6\textwidth]{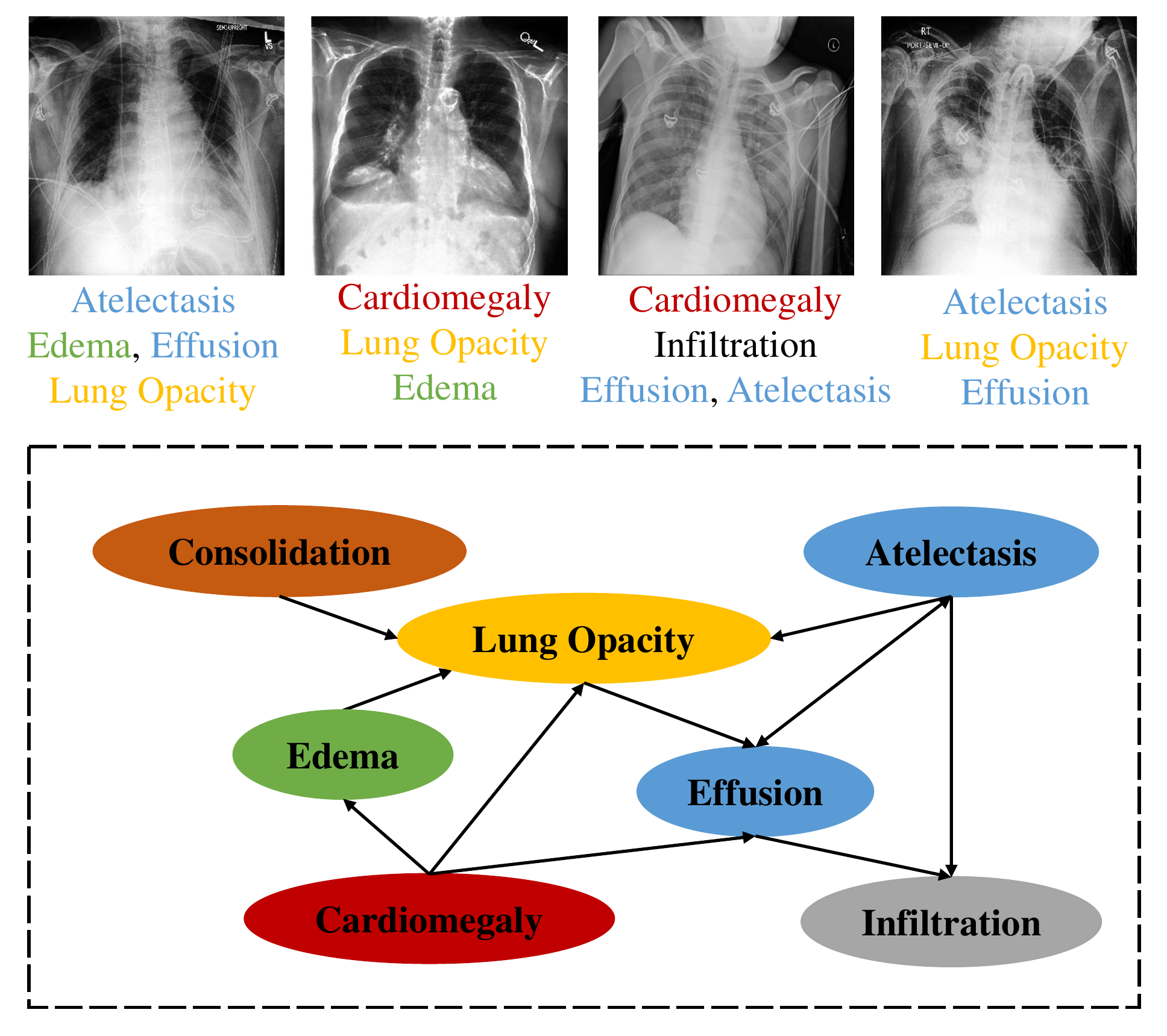} 
\caption{We constructed a directed graph over the pathological labels to model the global co-occurrence relationship in multi-label CXR recognition. For example, in the graph, "pathology A $\rightarrow$ pathology B" indicates that when pathology A appears, there is a great probability that pathology B has appeared, but the reverse may not hold.}
\label{fig:fig1}
\end{figure*}

\textit{Chest X-Ray} (CXR) imaging can effectively assist in the clinical management of chest diseases and is one of the most common screening techniques~\cite{hansell2008fleischner}.
The complex physiological relationships between chest diseases and the evaluation of thousands of radiology samples in a short period pose considerable diagnostic challenges for clinicians.
Therefore, multi-label CXR recognition is essential to assist clinicians in their diagnosis.
Compared to traditional image recognition tasks that predict only one label on each image, multi-label image recognition is more challenging. It requires more efficient methods to identify labels that appear simultaneously in an image.
Notably, researchers have found that some of the diseases in CXR are closely related. Moreover, clinical experience has demonstrated that co-occurrence relationships and correlations between pathological labels positively impact the final diagnosis. 
For example, atelectasis and effusion often lead to the development of infiltrates, and pulmonary edema often leads to cardiac hypertrophy~\cite{yao2017learning,wang2017chestx}. 
As shown in Fig.~\ref{fig:fig1}, exploring the co-occurrence relationship and interdependence of labels between diseases helps strengthen the intrinsic association of specific labels.
Therefore, the key to multi-label image recognition algorithms to assist CXR disease diagnosis is to model co-occurrence relationships and correlations between labels.

Early multi-label CXR recognition algorithms based on CNNs and their variants have made some progress by identifying each label in isolation and transforming the problem into multiple binary recognition tasks~\cite{yao2017learning,galleguillos2008object,shin2016learning}. 
However, they have not yet considered the correlation among labels and lacked modeling of the co-occurrence relationship among labels.
With the increase in research on label relations, the modeling of label correlations in a suitable way has become a subject of interest.
Some approaches rely on state-variable encoding and attentional mechanism-oriented modeling of local label information~\cite{clare2001knowledge,golfarelli2009survey,camnet,zhang2013review}. For example, Wang et al.~\cite{wang2016cnn} used RNN to model the relevance of labels in a sequential manner. The work of Zhu et al.~\cite{zhu2017learning} and Wang et al.~\cite{wang2017multi} through the attention mechanism considered limited local correlations among attentional regions of individual images. 
However, having complete and correct labels on every image is impractical.
Therefore, those approaches to modeling local labeling information are limited and unstable.
Furthermore, they ignored the global semantic topology among their labels, resulting in many unexplored relationships, such as global co-occurrence relationships, which are susceptible to noise from irrelevant and missing labels.

In recent years, Chen et al.~\cite{chen2019multi} proposed ML-GCN, and Chen et al.~\cite{chen2020label} proposed CheXGCN, both of which have achieved great success in multi-label image recognition tasks by capturing and learning relationships between labels based on label statistical information.
Similarly, the A-GCN designed by Li et al.~\cite{li2019learning} to capture label dependencies by constructing adaptive label structures has shown good performance. However, these methods all fuse image features with label embeddings through a simple linear model for multimodal feature fusion (e.g., simple summation, dot product, or concatenation) and have yet to consider the alignment between natural visual features in an image and label-relative embeddings. Also, they ignore the alignment between the distribution of labels on the image and the variation in the distribution of features present in the cross-modal features. Therefore, cross-modal fusion features obtained by this simple linear method can hardly be used effectively to express the complex correlations between different modal features fully.

Consider that previous approaches to modeling local label information have yet to be fully utilized to label global co-occurrence relationships. The approach to fusing features was carried out crudely, making it difficult to fuse different modal features effectively.
To this end, we propose a novel Bi-modal Bridged Graph Convolutional Network(BB-GCN) for multi-label CXR recognition, which mainly consists of a backbone module, a \textit{pathology Label Co-occurrence relationship Embedding} (LCE) module, and a \textit{Transformer Bridge Graph} (TBG) module.
In the Backbone module, we use the ViT~\cite{dosoViTskiy2020image} model to obtain a visual feature representation of each image.
In the LCE module, we use \textit{Graph Convolutional Network} (GCN)~\cite{kipf2016semi} to train and learn the pathology label co-occurrence relationship embedding that reflects the correlation relationship between different pathological labels. 
In the TBG module, we use the GroupSum method to bridge labeled symbiotic embeddings with image-level features to obtain a more compact and efficient fusion vector.
Finally, the multi-label loss function is used to optimize the framework to complete the end-to-end training.
We have developed the framework in two large-scale CXR datasets (ChestX-Ray14~\cite{wang2017chestx} and
CheXpert~\cite{irvin2019chexpert}) on which and evaluated the effectiveness of BB-GCN, achieving state-of-the-art performance on these: the mean AUC scores for the 14 pathologies were 0.813 and 0.835, respectively.
The contributions of our work including:
\begin{itemize}
\item Considering the limitations and instability of modeling local label information on a single modality, the \textit{pathology Label Co-occurrence relationship Embedding} (LCE) module is proposed to model the global co-occurrence relationships between multiple labels using graph structures.
\item The \textit{Transformer Bridge Graph} (TBG) module is proposed to bridge the cross-modal vectors more compactly and efficiently through the GroupSum method, considering the problems of feature misalignment and non-compactness when fusing different modal features. 
\item Experimental results on two multi-label CXR datasets show that our BB-GCN performs better than previous state-of-the-art models.
\end{itemize}

\section{Ralted work}
\label{sec:RelatedWorks}
In this section, we first review existing methods for multi-label image recognition on CXR datasets. Then, recent GCN-based studies are discussed, and several representative cross-modal fusion works are presented.
\subsection{ Multi-Label Chest X-Ray Recognition}
With the development of deep neural networks~\cite{krizhevsky2012imagenet, simonyan2014very, szegedy2015going, he2016deep, ibrahim2021deep,mahmud2020covxnet, ozturk2020automated}, multi-label CXR recognition has achieved great success on large-scale CXR datasets in the past few years.
In particular, datasets such as ChestX-Ray14 and CheXpert have been hot research topics for automatic CXR analysis.
Early approaches to multi-label image recognition naively divided the task into multiple independent binary recognition tasks that trained a set of classifiers for each label. Wang et al~\cite{wang2017chestx} used the classical AlexNet~\cite{krizhevsky2012imagenet}, VGGNet~\cite{simonyan2014very}, GoogleNet~\cite{ szegedy2015going}, ResNet~\cite{he2016deep} to evaluate the performance of multi-label CXR recognition, where ResNet-50~\cite{he2016deep} achieved the best results.
Rajpurkar et al.~\cite{rajpurkar2017chexnet} proposed CheXNet based on the DenseNet-121~\cite{huang2017densely} model with improvements.
Excellent performance was achieved in classifying abnormalities in each CXR and detecting pneumonia. Shen et al.~\cite{shen2018dynamic} designed a novel network training mechanism for efficiently training CNN-based automatic chest disease detection models. However, these approaches to modeling local label information ignore the relationship between pathology labels when classifying the models.

Researchers have begun to model local label information to capture correlations between pathological labels.
Yao et al.~\cite{yao2017learning}
achieved the correct recognition of multi-label CXR by redesigning the combined DenseNet and \textit{Long-Short Term Memory Network} (LSTM)~\cite{sak2014long}, demonstrating that the recognition performance of the model can be further improved by exploiting label correlation.
Furthermore, in terms of using attention mechanisms to establish the relationship between labels and regions of interest, Ypsilantis et al.~\cite{ypsilantis2017learning} used an RNN-based model to consecutively sample the entire CXR and concentrate on the most enlightening regions. Tang et al.~\cite{tang2018attention} designed an attention model to identify the disease and locate lesion regions based on the disease severity levels mined from radiology reports.
Guan et al.~\cite{guan2018diagnose} used CNNs to learn high-level features of images and designed attention-learning modules to provide additional attention guidance for chest disease recognition. In addition, Lee et al.~\cite{lee2018multi} also describe the relationship between labels by constructing a knowledge graph to obtain a more accurate image representation.
As mentioned above, those methods rely on state variable coding and attention mechanisms, model local label information, ignoring the global co-occurrence relationship between labels. 

\subsection{Graph Convolutional Network}
GCN was introduced in the literature~\cite{kipf2016semi} to perform semi-supervised recognition. The basic idea~\cite{camnet, jia2023end, hsgm, hpgn,sun2022multi} is to pass information between nodes and update the node representation through the graph adjacency matrix. Unlike standard convolution, which operates on local euclidean structures in images, GCN can model topologically structured data well. By learning the structural similarities between training data, GCN can map relationships to data features. 
As an effective technique for mining relationships, Shen et al.~\cite{hsgm} proposes a hierarchical similarity graph module to relieve the conflict of backbone networks and mine the discriminative features.
Marino et al.~\cite{marino2016more} employ graph neural networks to learn graph-structured prior knowledge and additional attribute relationships to assist visual recognition tasks effectively. 

In terms of multi-label zero-sample learning, Lee et al.~\cite{lee2018multi} design knowattention-learningscribe multiple correlations between labels and model the dependencies between labels through the learned information propagation mechanism. Wang et al.~\cite{wang2018zero} base their prediction of visual classifiers on the introduction of GCN using semantic embeddings and category relations. Based on the above GCN model, Yu et al.~\cite{yu2018modeling} used GCN and another neural network to model text and images, respectively, and trained by pairwise loss function to make the model performance significantly improved.
Shen et al.~\cite{hpgn} adopts a backbone network and a pyramidal graph network for vehicle retrieval.
Chen et al.~\cite{chen2019multi} proposed a practical framework that treats each label as a node and uses GCN to learn the relationship between labels for a classification task.
\subsection{Cross-modal Fusion}
Researchers often use concatenation or elemental summation to fuse different modal features better to achieve the fusion of cross-modal features. Fukui et al.~\cite{fukui2016multimodal} proposed that two vectors of different modalities are made outer products to achieve the fusion of multi-modal features by bilinear models. However, this method yields high-dimensional fusion vectors. 
Xu et al.~\cite{xu2021dual} encouraged data on both attribute and image modalities to be discriminated to improve attribute-image person re-identification.
To reduce the high-dimensional computation, Kim et al.~\cite{kim2016hadamard} designed the method, which achieves comparable performance to the work of Fukui et al. by doing Hadamard product between two feature vectors, but with slow convergence. It is worth mentioning that Zhou et al.~\cite{yu2018beyond} introduced a new method with stable performance and accelerated model convergence for the study of fusing image features and text embedding.
Inspired by the above research, we designed the \textit{Transformer Bridge Graph} (TBG) module based on bilinear pooling to bridge visual feature representation and pathology label co-occurrence relationship embedding. We achieved significant results on the multi-label CXR recognition task on two CXR datasets. Our model achieves learning of inter-label correlation through end-to-end training with significant performance compared to existing methods.
\begin{table}[!htp]
\centering
\caption{Preliminary symbols used in this paper.}
\label{tab:symbols}
\resizebox{\textwidth}{!}{%
\begin{tabular}{cl}
\hline
Symbols    & Explanation                                                   \\ \hline
$X$       & the sample images in the dataset                                        \\
$F$       & the image visual feature representation                       \\
$D$         & the number of dimensions of a vector               \\
$C$          & the number of pathology categories                            \\
$W$          & pathology word embeddings matrix                 \\
$L$       & the pathology label in the label set                            \\
$T$       & the occurrence times of the pathology                    \\
$A$          & pathology label correlation matrix             \\
$LO$          & pathology label co-occurrence relationship embeddings matrix \\

$l$          & the number of GCN layers                                      \\
$\Theta$          & the transformation-learnable parameters for each GCN layer    \\
$M$         & the input of the TBG module                    \\
$G$          & the number of groups in GroupSum operation              \\
$g$          & the number of units in each group $G$                           \\
$O$      & the predicted output of sample image                          \\
\hline
\end{tabular}%
}
\end{table}
\section{Materials and methods}\label{sec:BB-GCN}
\begin{figure}[!htp]
\centering
\includegraphics[width=\textwidth]{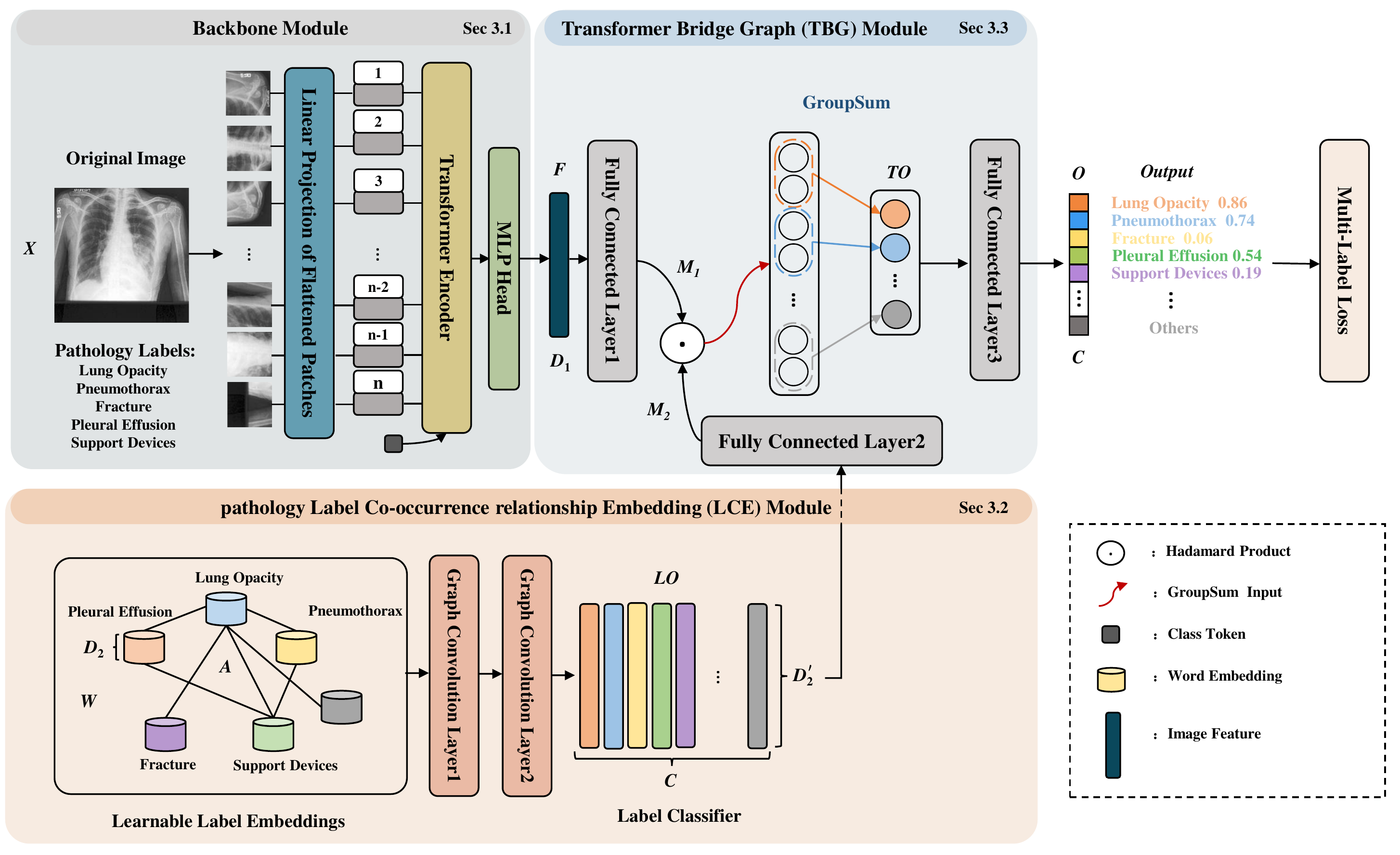} 
\caption{The overall framework of our proposed BB-GCN.}
\label{fig:model}
\end{figure}

Based on previous research, this section proposes a multi-label CXR recognition framework, BB-GCN, consisting of three main modules: the Backbone module, \textit{pathology Label Co-occurrence relationship Embedding} (LCE) module, and \textit{Transformer Bridge Graph} (TBG) bridging module. We first introduce the general framework of our model in Fig.~\ref{fig:model}, and then detail the workflow of each of these three modules. Lastly, we describe the datasets, implementation details, and evaluation metrics.

\subsection{Backbone Module}
ViT, with its excellent performance in various tasks, we use ViT-Base to accomplish the feature representation of images in this module precisely.
As shown in Fig.~\ref{fig:model} for this module, we first divide the image $X_i$ with 224$\times$224 resolution into $14$ patches of size 16$\times16$ by patch embeddings as input.
Second, to maximize the preservation of fine-grained feature information in the class token, the MLP-Head module last layer perceptron output in ViT-Base is adjusted to be the same as the class token, and this sub-network is then used to generate a $D_{1}$-dimensional representation $\bm{F_{i}}\in \mathbb{R}^{D_1}$ of each image $X_i$.

\subsection{Pathology Label Co-occurrence Relationship Embedding Module}
This section uses GCNs to learn pathology label co-occurrence relation embedding based on the relationships between different pathology labels. Unlike semi-supervised GCNs, where the node outputs are predicted scores, we design GCN nodes for learning the final outputs corresponding to the classifiers of the correlation pathology labels in the task. Specifically, our goal is to map pathological label dependencies in datasets based on GCN, to mark co-occurrence relationship embeddings between labels in a task.

As shown in Fig.~\ref{fig:model}, we use the GloVe~\cite{pennington2014glove} model to convert each pathology label (all $C$ kinds of pathology labels in the dataset) into an $D_{2}$-dimensional (i.e., 300-dimensional) word embedding. Thus, we can obtain the $C \times D_{2}$ pathology label word embeddings matrix $\boldsymbol{W}$. For example, there are 14 pathology categories in ChestX-Ray14, so one of the inputs to the first GCN layer is pathology label word embeddings matrix $\boldsymbol{W}$, which will be a $14\times300$ matrix.

In addition to obtaining the feature vectors for each node (pathology label), another basic problem is constructing the pathology label correlation matrix $\boldsymbol{A}$ between these nodes. In the implementation, we capture the label dependencies and construct the matrix $\boldsymbol{A}$ based on the label statistics on the whole dataset. Specifically, we count the number of occurrences ($T_i$) of the $i$-th pathology label ($L_i$) and the simultaneous occurrence times of $L_i$ and $L_j$ ($T_{ij} = T_{ji}$). In addition, the label relevance can be expressed by the conditional probability as follows.
\begin{equation}\label{equ:shizi2}
    P_{ij}=P \left(L_i|L_j \right)= \frac{T_{ij}}{T_j},\forall i \in \left[1,C\right], 
\end{equation}
where $\boldsymbol{P}_{ij}$ denotes the probability that $L_i$ occurs under the condition that $L_j$ occurs.
Note that since the conditional probabilities between two objects are asymmetric, $\bm{P}_{ij} \neq \bm{P}_{ji}$.

We can construct the pathology label correlation matrix $\boldsymbol{A}$ on this basis, and the element value $\boldsymbol{A}_{ij}$ at each position in this matrix is equal to $\boldsymbol{P}_{ij}$.
However, if we trained directly using this immature correlation matrix, the rare co-occurring pathological inter-label relationships would become some noise affecting the data distribution and the model convergence. To filter the noise, we binarize the above label correlation matrix $\boldsymbol{A}$ and set the threshold $\epsilon$ for updating:
\begin{equation}\label{equ:shizi4}
    \bm{A}_{i j}= \begin{cases}0, & \text { if } \bm{P}_{i j} \leq \epsilon \\ 1, & \text { otherwise }\end{cases},
\end{equation}
where $\epsilon \in \left[ 0,1 \right]$. In addition, when using GCN to update node features during information propagation, the binary correlation matrix causes the GCN to produce an over-smoothing problem, and the generated node features are not distinguishable. We use a weighting scheme to calculate the final enhanced pathology label correlation matrix $\boldsymbol{EA}$ as:
\begin{equation}\label{equ:shizi5}
    \bm{EA}_{i j}= \begin{cases}\frac{\delta}{\sum_{i=1 \cap i \neq j}^{C}\bm{A}_{i j}} \bm{A}_{i j}, & \text { if } i \neq j \\ 1-\delta, & \text { otherwise }\end{cases},
\end{equation}
where $\delta \in \left[ 0,1 \right]$, and $\delta$ determines the weights assigned to a node itself and other related nodes. By doing this, we can have a fixed weight for the node itself when updating the node features and weight the related nodes as determined by the neighborhood distribution. When $\delta \rightarrow$ 1, the features of the node itself will not be considered. On the other hand, when $\delta \rightarrow 0$, the information of the relevant neighbors will often be ignored. In this way, we can use this enhanced label correlation matrix $\boldsymbol{EA}$ to update the features of the nodes by choosing the appropriate $\delta$.

After obtaining the pathology label word embeddings matrix $\boldsymbol{W}$ and the enhanced pathology label correlation matrix $\boldsymbol{EA}$, we design a two-layer GCN to propagate the information, and each GCN layer can be described as:
\begin{equation}\label{equ:shizi7}
    \bm{H^{l+1}} = f^{l} \left( \bm{\widetilde{EA}H^l \Theta^l} \right),  l \in \left[0,2\right]
\end{equation}
$f(\cdot)$ denotes a non-linear operation, which is acted by LeakyReLU~\cite{maas2013rectifier} with scope 0.2 in our experiments.
$\boldsymbol{\widetilde{EA}}$ (see~\cite{kipf2016semi} for details) denotes the normalized version of the enhanced label correlation matrix $\boldsymbol{EA}$.
Where $\boldsymbol{H^{l}}$ is the feature descriptions of $l$-th GCN layer, and $\boldsymbol{\Theta^{l}}$ is the corresponding trainable transformation matrix. 

Note that $\boldsymbol{H}$ is the input of this sub-network, which contains pathology label word embeddings matrix $\boldsymbol{W}$ and the enhanced pathology label correlation matrix $\boldsymbol{EA}$. When $l$=2, the output of LCE module is $\bm{LO}$=$\bm{H^{3}}$, which is the $C\times \acute{D_{2}}$ (i.e., 768-dimensional) pathology label co-occurrence relationship embeddings matrix $\boldsymbol{LO}$. Each row of $\boldsymbol{LO}$ corresponds to the label classifier of the corresponding label in the task and will be bridged with the visual feature representation vector $\boldsymbol{F_i}$ in the TBG module.

\subsection{Transformer Bridge Graph Module}
Unlike previous work that fused cross-modal vectors using dot product.
In this section, we design the TBG module to efficiently bridge the visual feature representation and pathology label co-occurrence relationship embeddings based on the bilinear pooling~\cite{lin2015bilinear} approach, which helps to achieve higher performance of BB-GCN.

As shown in Fig.~\ref{fig:model} of this module, on the one hand, the Hadamard product increases the interaction between different modal vectors, thus improving the accuracy of BB-GCN.
On the other hand, GroupSum operation reduces overfitting and parameter explosion by group mapping the elements after the Hadamard product, which speeds up the convergence of the BB-GCN. The input of the TBG module consists of two parts: the visual image feature $\boldsymbol{F}$ and pathology label co-occurrence relationship embedding matrix $\boldsymbol{LO}$. 
Formally, given the visual feature representation $\boldsymbol{F_i}$ of the $i$-th image, for $ {\forall}j \in \left[ 1,C \right]$, we bridge $\boldsymbol{F_i}$ and $\boldsymbol{LO_j}$ to generate the $j$-th element of final predicted output $O_j$. First, we use two FC layers to convert $\boldsymbol{F_i}$ into an $D_3$-dimensional vector $\boldsymbol{M_1}$ and $\boldsymbol{LO_j}$ into an $D_3$-dimensional vector $\boldsymbol{M_2}$, respectively:
\begin{equation}\label{equ:shizi8}
    \left\{
    \begin{aligned}
        & \bm{M_1}=FC_{1}(\bm{F_i})\in\mathbb{R}^{D_{3}}\\
         & \bm{M_2}=FC_{2}(\bm{LO_j})\in\mathbb{R}^{D_{3}} 
    \end{aligned},
    \right.
\end{equation}
Second, the initial bilinear pooling on which the TBG module is based is defined as follows:
\begin{equation}\label{equ:shizi9}
    \bm{TO} =\bm{M^{T}_{1}S_{i}M_2},
\end{equation}
where $\bm{TO}\in \mathbb{R}^{o}$ is the output of the TBG module, $ \bm{S_{i}} \in \mathbb{R}^{ D_{3}\times D_{3} } $ is the bilinear mapping matrix with bias terms included.
$ \bm{S}=\left[S_i,\cdots,S_o\right]\in \mathbb{R}^{D_{3} \times D_{3} \times o}$ can be decomposed into two low-rank matrices $\bm{u_{i}}=\left[u_1,\cdots,u_G\right]\in \mathbb{R}^{D_{3} \times G}$,
$\bm{v_{i}}=\left[v_1,\cdots,v_G\right]\in \mathbb{R}^{D_{3} \times G}$. Therefore, the Eq.(\ref{equ:shizi9}) can be rewritten as follows:

\begin{equation}\label{equ:shizi10}
    \begin{aligned}         
    \bm{TO_{i}}=\bm{1^{T}} \left(\bm{u_{i}^{T}M_{1}} \circ \bm{v_{i}^{T}M_{2}} \right)
    \end{aligned}
\end{equation}
where the value of $G$ is the factor or latent dimension of two low-rank matrices
, $1^{T}\in \mathbb{R}^{G}$ is an all-one vector. In order to obtain the final $\boldsymbol{TO}$, two three-dimensional tensors 
$\bm{u_{i}}\in \mathbb{R}^{D_{3} \times G \times o},
\bm{v_{i}}\in \mathbb{R}^{D_{3} \times G \times o}$ need to be learned.
Under the premise of ensuring the generality of Eq.(\ref{equ:shizi10}), 
the two learnable tensors $\boldsymbol{u}, \boldsymbol{v}$ are converted into two-dimensional matrices by matrix variable dimension, namely $\bm{u_{i}} \rightarrow \bm{\tilde{u}}\in \mathbb{R}^{D_{3}\times Go}$ and $\bm{v_{i}} \rightarrow \bm{\tilde{v}}\in \mathbb{R}^{D_{3}\times Go}$, Eq.(\ref{equ:shizi10}) simplifies to:
\begin{equation}\label{equ:shizi11}
    \begin{aligned}         
            \bm{TO}= GroupSum \left( \bm{\tilde{u}^{T}M_{1}} \circ \bm{\tilde{v}^{T}M_{2}} , G \right)
    \end{aligned}
\end{equation}
where the function \textit{GroupSum}$\left(\bm{vector}, G \right)$
represents the mapping of $g$ elements in $\boldsymbol{vector}$ into $\frac {1}{G}$ groups and outputting all $G$ groups obtained after complete mapping as potential dimensions, $\boldsymbol{TO} \in \mathbb{R}^{G}$.

Finally, we employ an FC layer to generate the $j$-th element of the final predicted output:
\begin{equation}\label{equ:shizi12}
    \begin{aligned}         
        O_j = FC_{3}\left ( \bm{TO}\right )
    \end{aligned}
\end{equation}

After visual feature representation $\boldsymbol{F_i}$ is bridged $C$ times with pathology label co-occurrence relationship embedding $\boldsymbol{LO_{j}}$, we will get the complete predicted output $O$ corresponding to $\boldsymbol{F_i}$.
In addition, all FC layers share parameters at each bridging.
Finally, we update the entire network end-to-end using the MultiLabelSoftMargin Loss (Multi-labelLoss).
The training loss function is described as:
\begin{equation}\label{equ:shizi13}
    \begin{aligned}
    \mathcal{L}\left(O, L\right)=&-\frac{1}{C} \sum_{j=1}^{C} L_{j} \log \left(\left(1+\exp \left(-O_{j}\right)\right)^{-1}\right)
    \\&+\left(1-L_{j}\right) \log \left(\frac{\exp \left(-O_{j}\right)}{\left(1+\exp \left(-O_{j}\right)\right)}\right),
    \end{aligned}
\end{equation}
where $O$ and $L$ denote the predicted label and the true label of the sample image, respectively.
$O_{j}$ and $L_{j}$ denote the $j$-th element in its predicted label and the $j$-th element in the actual label.

\subsection{Datasets}
ChestX-Ray14 is a large-scale CXR dataset published by the National Institutes of Health that
consists of 112,120 frontal view images. Containing 78,466 training images, 11,220 validation images, and 22,434 test images.
The patient images are labeled with an average of about 1.6 pathology labels from 14 semantic categories.
To make our results directly comparable to most published baselines, we strictly follow the official split standards of ChestX-Ray14 provided by Wang et al.~\cite{wang2017chestx} to conduct our experiments.
We use the train and valid sets to train our model and then evaluate the performance on the test set.

CheXpert is a popular multi-label CXR dataset for the identification, detection, and segmentation of common diseases of the chest and lungs.
It consists of 224,316 images, 12 disease symptom labels, and two non-disease labels (No Finding and Support Device).
Each image is assigned one or more disease symptoms, and the results of the diseases are labeled as positive, negative, and indeterminate; if no positive disease is found in the image, it is labeled as "No Finding".
Indeterminate labels in the images can be considered positive
(CheXpert$\_$1s) or negative (CheXpert$\_$0s).
On average, each image was labeled with 2.9 pathology labels in chexpert$\_$1s and 2.3 pathology labels in CheXpert$\_$0s.
Since the data for the test set is not yet publicly available, to better fit this task.
We re-divided the dataset into a training set, validation set, and test set in the ratio of 7:1:2.

\begin{figure}[htbp]
\centering
\includegraphics[width=\textwidth]{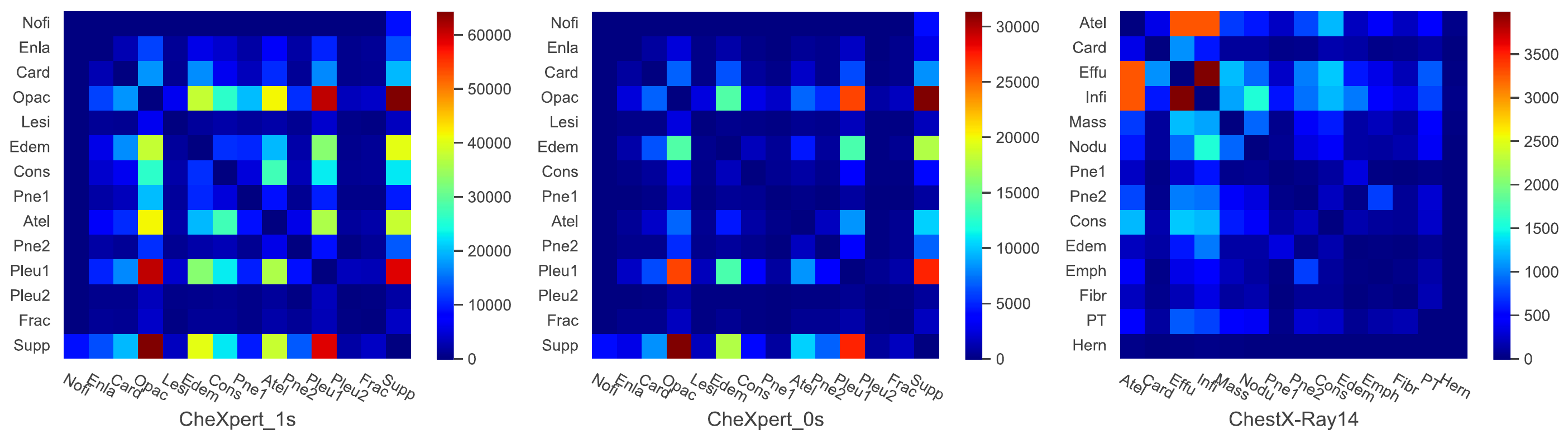} 
\caption{Graphical representation of the pathology label co-occurrence matrix extracted from the CheXpert\_1s, CheXpert\_0s and ChestX-Ray14 datasets, respectively.}
\label{fig:fig3}
\end{figure}

As previously described, the proposed LCE module involves global modeling of all pathology labels based on co-occurring pairs—the results in identifying potential pathologies present in each image. 
As shown in Fig.~\ref{fig:fig3}, many pathology label pairs with co-occurrence relationships were obtained by counting the occurrences of all pathology labels in both datasets separately. For example, pulmonary disease is often associated with pleural effusion, and atelectasis is often co-occurring with infiltration. This phenomenon provides preliminary evidence of the feasibility of the proposed LCE module.

\subsection{Implementation details}
All experiments were run on Intel 8268 CPU and NVIDIA Tesla V100 32GB GPU.
Moreover, implemented based on the PyTorch framework.
First, we resize all images to 256$\times$256 and normalize via the mean and standard deviation of the ImageNet dataset~\cite{deng2009imagenet}.
Then, apply the random crop to make images 224$\times$224, random horizontal flip, and random rotation, as some images may be ﬂipped or rotated within the dataset.
The output characteristic dimension $D_{1}$ of the backbone is 768.
In the LCE module, our BB-GCN consists of two layers of GCNs by stacking, with relationship dimensions of 1024 and 768.
The input pathology label word embedding is a 300-dimensional vector generated by the GloVe~\cite{pennington2014glove} model pre-trained on the Wikipedia dataset.
When multiple words represent the pathology labels, we use the average vector of all words as the pathology label word embedding.
In constructing the correlation matrix, we determined by subsequent ablation experiments to set $\epsilon$=0.3 in Eq.(\ref{equ:shizi4}), respectively.
In Eq.(\ref{equ:shizi5}), set $\delta$=0.2.
In the TBG bridging module, we set $D_{3}$=384 to bridge the vectors of the two modes.
Furthermore, set $G$=64 with $g$=6 to complete the GroupSum method.
The whole network is updated by \textit{Stochastic Gradient Descent} (SGD) with a momentum of 0.9 and a weight decay of 5E-5.
The initial learning rate of the backbone module and the TBG module is 0.001, while the initial learning rate of the LCE module is 0.01.
All learning rates decay ten times every 10 epochs.
\subsection{Evaluation criteria}
In our experiments, we used the AUC score for each pathology and the mean AUC score across all cases to measure the performance of the BB-GCN.
There is no data overlap between the training and testing subsets.
The true label of each image was labeled with $L= \left[L_1, L_2, \dots, L_C \right]$.
In the dataset of two CXR label numbers $C$=14, each element $L_C$ indicates the presence or absence of the $C$-th pathology, i.e., 1 indicates presence, and 0 indicates absence.
For each image, the label was predicted as positive if the confidence level of the label was greater than 0.5. For a fair comparison, we also calculated and reported the mean \textit{Overall Precision} (OP), \textit{Recall} (OR), \textit{F1} (OF1), and other traditional image recognition metrics for performance evaluation, as shown in the Eq.(\ref{equ:shizi14}).
\begin{equation}\label{equ:shizi14}
    \begin{aligned}
    OR=\frac{\sum_i n_i^c}{\sum_i n_i^g} , OP=\frac{\sum_i n_i^c}{\sum_i n_i^p} , OF1=\frac{2 \times OP \times OR}{OP + OR}
    \end{aligned},
\end{equation}
where $n^{c}_{i}$ represents the number of images correctly predicted for the $i$-th label, $n^{g}_{i}$ represents the number of images actually annotated for the $i$-th label, and $n^{p}_{i}$ represents the number of all images predicted for the $i$-th label.

\section{Experiment results and discussion}
\label{Experiment}
In this section, we report and discuss results on two benchmark multi-label CXR recognition datasets. Ablation experiments were also conducted to explore the effects of different parameters and components on BB-GCN, and finally, visual analysis was performed.

\subsection{Comparison with existing methods}
In this section, we conducted experiments on ChestX-Ray14 and CheXpert, respectively, to compare the performance of BB-GCN with existing methods.

\begin{table*}[!htp]
\centering
\caption{AUC comparisons of BB-GCN with existing methods on ChestX-Ray14.}
\label{tab:NIH_sota}
\resizebox{\textwidth}{!}{%
\begin{tabular}{ccccccccccccccccc}
\hline
                  & \multicolumn{14}{c}{ChestX-Ray14}                                                                                                                                                                                &                      \\ \cline{2-15} 
Method            & \multicolumn{14}{c}{AUC}                                                                                                                                                                                         & Mean AUC             \\ \cline{2-15}
              & Atel & Card & Effu & Infi & Mass & Nodu & Pne1 & Pne2 & \multicolumn{1}{l}{Cons} & \multicolumn{1}{l}{Edem} & \multicolumn{1}{l}{Emph} & \multicolumn{1}{l}{Fibr} & \multicolumn{1}{l}{PT} & \multicolumn{1}{l}{Hern} & \multicolumn{1}{l}{} \\ \hline
U-DCNN~\cite{wang2017chestx}        & 0.700   & 0.810   & 0.759   & 0.661   & 0.693   & 0.669   & 0.658   & 0.799   & 0.703                       & 0.805             & 0.833                       & 0.786                       & 0.684                     & 0.872                      & 0.745                   \\
LSTM-Net~\cite{yao2017learning}         & 0.733   & 0.856   & 0.806   & 0.673   & 0.718   & 0.777   & 0.684   & 0.805   & 0.711                       & 0.806          & 0.842                       & 0.743                       & 0.724                     & 0.775                       & 0.761                   \\
DR-DNN~\cite{shen2018dynamic}             & 0.766   & 0.801   & 0.797   & 0.751   & 0.760   & 0.741   & \pmb{0.778}   & 0.800   & 0.787                       & 0.820                       & 0.773                       & 0.765                       & 0.759                      &  0.748                     & 0.775                   \\
AGCL~\cite{tang2018attention}              & 0.756   & 0.887   & 0.819   & 0.689   & 0.814   & 0.755   & 0.729   & 0.850   & 0.728                       & 0.848                       & 0.906                       & 0.818                       & 0.765                     & 0.875                       & 0.803                   \\
CheXNet~\cite{rajpurkar2017chexnet}             & 0.769   & 0.885   & 0.825   & 0.694   & 0.824   & 0.759   & 0.715   & 0.852   & 0.745                       & 0.842                       & 0.906                       & 0.821                       & 0.766                     & 0.901                       & 0.807                   \\
DNet~\cite{guendel2018learning}             & 0.767   & 0.883   & 0.828   & 0.709   & 0.821   & 0.758   & 0.731   & 0.846   & 0.745                       & 0.835                       & 0.895                       & 0.818                       & 0.761                     & 0.896                       & 0.807                   \\
CRAL~\cite{guan2018diagnose}              & 0.781   & 0.880   & 0.829   & 0.702   & 0.834   & 0.773   & 0.729   & 0.857   & 0.754                       & 0.850                       & 0.908                       & 0.830                       & 0.778                     & 0.917                       & 0.816                   \\
DualCheXN~\cite{chen2019dualchexnet}              & 0.784   & 0.888   & 0.831   & 0.705   & 0.838   & 0.796   & 0.727   & 0.876   & 0.746                       & 0.852                       & 0.942                       & \pmb{0.837}                       & \pmb{0.796}                     & 0.912                       & 0.823                   \\
CheXGCN~\cite{chen2020label}              & 0.786   & 0.893   & 0.832   & 0.699   & 0.840   & \pmb{0.800}   & 0.739   & 0.876   & 0.751                       & 0.850                       & \pmb{0.944}                       & 0.834                       & 0.795                     & \pmb{0.929}                       & 0.826                   \\ 
\textbf{BB-GCN(Ours)}         & \pmb{0.859}   & \pmb{0.908}   & \pmb{0.922}   & \pmb{0.773}   & \pmb{0.880}   & 0.784   & 0.768   & \pmb{0.915}   & \pmb{0.832}                       & \pmb{0.910}                       & 0.897                       & 0.737                       & \pmb{0.796}                     & 0.715                       & \pmb{0.835}                   \\ 
\hline         
\end{tabular}%
}
\tiny{
{$^{\ast}$The 14 pathologies in Chest X-Ray14 are Atelectasis (Atel), Cardiomegaly (Card), Effusion (Effu), Infiltration (Inﬁ), Mass, Nodule (Nodu), Pneumonia (Pne1), Pneumothorax (Pne2), 
Consolidation (Cons), Edema (Edem), Emphysema (Emph), Fibrosis (Fibr), Pleural Thickening (PT)  and Hernia (Hern), respectively.}}
\end{table*}
\subsubsection{Results from ChestX-Ray14 and discussion}
We compared BB-GCN with a variety of existing methods
including U-DCNN~\cite{wang2017chestx},
LSTM-Net~\cite{yao2017learning},
CheXNet~\cite{rajpurkar2017chexnet},
DNet~\cite{guendel2018learning},
AGCL~\cite{tang2018attention},
DR-DNN~\cite{shen2018dynamic},
CRAL~\cite{guan2018diagnose},
DualCheXN~\cite{chen2019dualchexnet},
and CheXGCN~\cite{chen2020label}.
We present the results of the comparison on ChestX-Ray14 in Table~\ref{tab:NIH_sota}
including the evaluation metrics for the entire dataset of 14 pathology labels. 
BB-GCN outperformed all candidate methods on most pathology-labeled metrics.
Fig.~\ref{fig:roc} illustrates the ROC curves of our model over the 14 pathologies on ChestX-Ray14.
Specifically, BB-GCN outperforms these previous methods in mean AUC score, especially for U-DCNN (0.745) and LSTM-DNet (0.761), with improvements of 9.0\% and 7.4\%. 
Moreover, it outperforms DualCheXNet (0.823) and improves the AUC score of detecting Mass (0.880 vs. 0.838) and Pneumonia (0.768 vs. 0.727) by more than 4.0\%. 
Notably, the mean AUC score of BB-GCN has an improvement of 0.9\% over CheXGCN (0.826). The AUC scores of some pathologies labeled with BB-GCN are obviously improved, e.g., Infiltration (0.773 vs. 0.699), Consolidation (0.832 vs. 0.751), Edema (0.910 vs. 0.850), and Atelecta (0.859 vs. 0.786). 
It has to be mentioned that our proposed model performs somewhat poorly on Nodule, Fibrosis, and Hernia labels.
Note that the pathogenesis of these diseases is systemic, and we generated word embeddings of their pathological labels using only their noun meanings without adding additional semantics to explain their sites of pathogenesis.
This reason led to the unsatisfactory performance of BB-GCN on these labels.
Overall, the proposed BB-GCN improves the overall multi-label recognition performance of ChestX-Ray14 and outperforms existing methods.
\begin{table*}[!htp]
\centering
\caption{AUC comparisons of BB-GCN with previous baseline on CheXpert\_1s.}
\label{tab:1s_sota}
\resizebox{\textwidth}{!}{%
\begin{tabular}{cccccccccccccccc}
\hline
                  & \multicolumn{14}{c}{CheXpert\_1s}                                                                                                                                                                                &                      \\ \cline{2-15} 
Method            & \multicolumn{14}{c}{AUC}                                                                                                                                                                                         & Mean AUC             \\ \cline{2-15}
                  & Nofi    & Enla  & Card  & Opac  & Lesi  & Edem  & Cons  & Pne1  & Atel  & \multicolumn{1}{l}{Pne2} & \multicolumn{1}{l}{Pleu1} & \multicolumn{1}{l}{Pleu2} & \multicolumn{1}{l}{Frac} & \multicolumn{1}{l}{Supp} & \multicolumn{1}{l}{} \\ \hline
ML-GCN~\cite{chen2019multi}            & 0.879   & 0.630 & 0.841 & 0.723 & 0.773 & 0.856 & 0.692 & 0.740 & 0.713 & 0.829                    & 0.873                     & 0.802                     & 0.762                    & 0.868                    & 0.784                \\
U\_Ones~\cite{irvin2019chexpert}                  & 0.890      & 0.659    & 0.856    & 0.735    & 0.778    & 0.847    & 0.701    & 0.756    & \pmb{0.722}    & 0.855                       & 0.871                        & 0.798                        & 0.789                       & \pmb{0.878}                       & 0.795                   \\
ViT-Base~\cite{dosoViTskiy2020image}          & 0.860   & 0.649 & 0.828 & 0.721 & 0.773 & 0.811 & 0.694 & 0.754 & 0.686 & 0.804                    & 0.859                     & 0.769                     & 0.758                    & 0.837                    & 0.772                \\
\textbf{BB-GCN\_1s (Ours)}         & \pmb{0.891}   & \pmb{0.708}   & \pmb{0.868}   & \pmb{0.740}   & \pmb{0.788}   & \pmb{0.857}   & \pmb{0.754}   & \pmb{0.770}   & 0.707   & \pmb{0.875}                       & \pmb{0.879}                        & \pmb{0.843}                        & \pmb{0.793}                       & 0.866                       & \pmb{0.810} \\ \hline
\end{tabular}%
}
\tiny{
{$^{\ast}$The 14 pathologies in CheXpert are No Finding (Nofi), Enlarged Cardiomediastinum (Enla), Cardiomegaly (Card), Lung Opacity (Opac), Lung Lesion (Lesi), Edema (Edem), Consolidation (Cons), Pneumonia (Pne1), Atelectasis (Atel), Pneumothorax (Pne2), Pleural Effusion (Pleu1), Pleural Other (Pleu2), Fracture (Frac) and Support Devices (Supp).
}}
\end{table*}
\begin{table*}[!htp]
\centering
\caption{AUC comparisons of BB-GCN with the previous baseline on CheXpert\_0s.}
\label{tab:0s_sota}
\resizebox{\textwidth}{!}{%
\begin{tabular}{cccccccccccccccc}
\hline
                  & \multicolumn{14}{c}{CheXpert\_0s}                                                                                                                                                                                &                      \\ \cline{2-15} 
Method            & \multicolumn{14}{c}{AUC}                                                                                                                                                                                         & Mean AUC             \\ \cline{2-15}
              & Nofi & Enla & Card & Opac & Lesi & Edem & Cons & Pne1 & Atel & \multicolumn{1}{l}{Pne2} & \multicolumn{1}{l}{Pleu1} & \multicolumn{1}{l}{Pleu2} & \multicolumn{1}{l}{Frac} & \multicolumn{1}{l}{Supp} & \multicolumn{1}{l}{} \\ \hline
 
 ML-GCN~\cite{chen2019multi}             & 0.864   & 0.673   & 0.831   & 0.681   & 0.802   & 0.770   & 0.713   & 0.758   & 0.654   & 0.845                       & 0.841                        & 0.764                        & 0.754                       & 0.838                       & 0.771                   \\ 
 U\_Zeros~\cite{irvin2019chexpert}             & 0.885   & 0.678   & 0.865   & 0.730   & 0.760   & 0.853   & \pmb{0.735}   & 0.740   & 0.700   & \pmb{0.872}                       & \pmb{0.880}                        & 0.775                        & 0.743                       & 0.877                       & 0.792                   \\
 ViT-Base~\cite{dosoViTskiy2020image}             & 0.871   & 0.679   & 0.848   & 0.694   & 0.815   & 0.787   & 0.722   & 0.772    & 0.656  & 0.846   & 0.840                       & 0.799                        & 0.795                        & 0.821                       & 0.782                                     \\   
\textbf{BB-GCN\_0s (Ours)}         & \pmb{0.900} & \pmb{0.681} & \pmb{0.866} & \pmb{0.744} & \pmb{0.823} & \pmb{0.858} & 0.720 & \pmb{0.775} & \pmb{0.732} & 0.863                    & 0.877                     & \pmb{0.846}                     & \pmb{0.820}                    & \pmb{0.878}                   & \pmb{0.813}                \\ \hline
\end{tabular}%
}
\end{table*}
\subsubsection{Results from CheXpert and discussion}
To our limited knowledge, CheXpert is a new dataset, and its test set has yet to be publicly available and can only be re-divided by itself. Fewer state-of-the-art methods are available for comparison.
Based on that, we further evaluated the comparison of our model with the uncertainty labeling treatments mentioned in the original dataset (U\_Ones and U\_Zeros)~\cite{irvin2019chexpert}. 
As shown in Table~\ref{tab:1s_sota}, BB-GCN\_1s obtained higher mean AUC scores on 14 pathological labels for CheXpert\_1s, which were 1.5\% higher than the techniques in the original paper U\_Ones. 
Also, compared to the vanilla ViT-Base, the improvement is 3.8\%.
As shown in Table~\ref{tab:0s_sota}, BB-GCN\_0s obtained higher mean AUC scores on 14 pathological labels for CheXpert\_0s, which were 2.1\% higher than the techniques U\_Zeros in the original paper.
The mean AUC score of BB-GCN is 3.1\% higher than vanilla ViT-Base. Those results prove that our two proposed modules can work better when reinforcing each other.

\begin{figure}[!htbp]
  \centering
  \includegraphics[width=\textwidth]{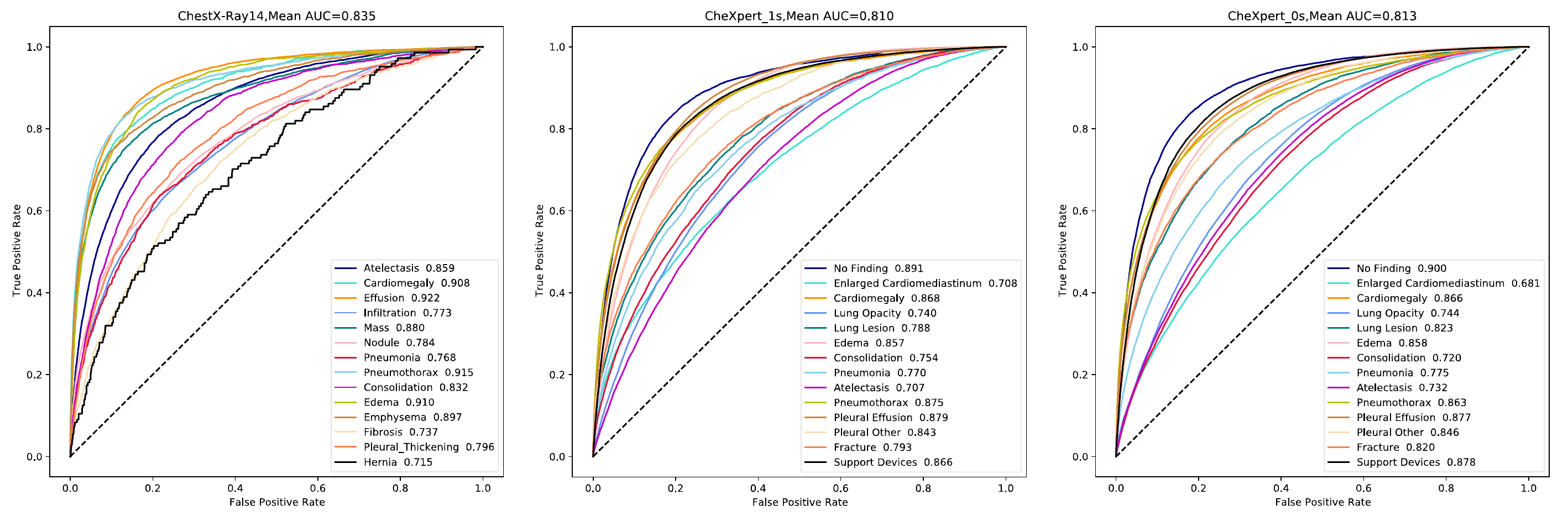} 
  \caption{ROC curves of BB-GCN on the ChestXRay14, and CheXpert, respectively. The corresponding AUC scores are given in Table~\ref{tab:NIH_sota}, Table~\ref{tab:1s_sota} and Table~\ref{tab:0s_sota}.}
  \label{fig:roc}
\end{figure}

Overall, the AUC score of BB-GCN\_1s was better than BB-GCN\_0s by 0.3\%, especially in Lung Lesion by 3.5\% (0.788$\rightarrow$0.823) and in Atelectasis by 2.5\% (0.707$\rightarrow$0.732), and 2.7\% (0.793$\rightarrow$0.820) on Fracture. This is because the true value of these uncertainty labels on the image is likely to be negative. The converse is also true.
Fig.~\ref{fig:roc} illustrates the ROC curves of BB-GCN on ChestX-Ray14, CheXpert\_1s, and CheXpert\_0s for the 14 pathologies, respectively.
\subsection{Ablation experiments and discussion}
\subsubsection{BB-GCN with its different components on ChestX-Ray14}
We have experimented with the performance of the components of the BB-GCN, and the results are shown in Table~\ref{tab:table-xiaorongshiyan}.
Compared to the baseline ViT-Base, the mean AUC score of Baseline+LCE was significantly higher by 2.8\% (0.781 $\rightarrow$ 0.809), 
especially in Atelectasis (0.759 $\rightarrow$ 0.817), Cardiomegaly (0.767 $\rightarrow$ 0.905), Effusion (0.838 $\rightarrow$ 0.888), Infiltration(0.720 $\rightarrow$ 0.735) on pathology, exceeded the vanilla ViT-Base by an average of 6.5\% in those pathology labels. 
The proposed LCE module is crucial in mining the global co-occurrence between pathology labels. 
Compared to the ViT-Base baseline, the mean AUC score of Baseline+TBG was significantly higher by 3.0\% (0.781 $\rightarrow$ 0.811), 
especially in Atelectasis (0.759 $\rightarrow$ 0.830), Effusion (0.838 $\rightarrow$ 0.906), Pneumothorax (0.799 $\rightarrow$ 0.895), Mass (0.711 $\rightarrow$ 0.856) on pathology, exceeded the vanilla ViT-Base by an average of 8.5\% in those pathology labels.
\begin{table*}[!htp]
\centering
\caption{Comparison of AUC of BB-GCN with its different components on ChestX-Ray14.}
\label{tab:table-xiaorongshiyan}
\resizebox{\textwidth}{!}{%
\begin{tabular}{cccccccccccccccc}
\hline
\multicolumn{1}{l}{}         & \multicolumn{14}{c}{Chest X-Ray14}                                                            &          \\ \cline{2-15}
Method    & \multicolumn{14}{c}{AUC}                                                                      & Mean AUC \\ \cline{2-15}
         & Atel & Card & Effu & Infi & Mass & Nodu & Pneu1 & Pneu2 & Cons & Edem & Emph & Fibr & PT & Hern &          \\ \hline
\multicolumn{1}{l}{Baseline~\cite{dosoViTskiy2020image}}  & 0.759   & 0.767   & 0.838   & 0.720   & 0.711   & 0.705   & 0.745   & 0.799   & 0.810                       & 0.902                       & 0.795                       & \pmb{0.776}                       & 0.706                     & \pmb{0.903}                       & 0.781                   \\
\multicolumn{1}{l}{Baseline+LCE}  & 0.817   & 0.905   & 0.888   & 0.735   & 0.764  & 0.759   & 0.736   & 0.853   & 0.828                       & 0.906                       & 0.854                       & 0.724                       & 0.779                     & 0.784                       & 0.809                  \\
\multicolumn{1}{l}{Baseline+TBG}  & 0.830   & 0.882   & 0.906   & 0.747   & 0.856  & 0.756   & 0.752   & 0.895   & 0.815                       & 0.898                       & 0.877                       & 0.707                       & 0.780                     & 0.660                       & 0.811                  \\
\multicolumn{1}{l}{\textbf{BB-GCN (Ours)}}        & \pmb{0.859}   & \pmb{0.908}   & \pmb{0.922}   & \pmb{0.773}   & \pmb{0.880}   & \pmb{0.784}   & \pmb{0.768}   & \pmb{0.915}   & \pmb{0.832}                       & \pmb{0.910}                       & \pmb{0.897}                       & 0.737                       & \pmb{0.796}                     & 0.715                       & \pmb{0.835}                   \\ \hline         
\end{tabular}%
}
\end{table*}

Note that the direct input to the TBG module in the Baseline+TBG model is a vector of 14 pathology-labeled words with initial semantic information. The experimental results demonstrate our proposed TBG module of superior cross-modal vector bridging capability.
With the addition of LCE module and TBG module, the BB-GCN significantly improved the mean AUC score by 2.6\%. In particular, the AUC score improvement was significant for Atelectasis (0.817 $\rightarrow$ 0.859), Pneumothorax (0.853 $\rightarrow$ 0.915), and Emphysema (0.854 $\rightarrow$ 0.897). This phenomenon indicates that LCE and TBG modules in our framework can reinforce and complement each other to make the BB-GCN perform at its best.
\subsubsection{The $\epsilon$ and $\delta$ in the pathology label correlation matrix}
\begin{figure}[!htpb]
  \centering
  \includegraphics[width=\textwidth]{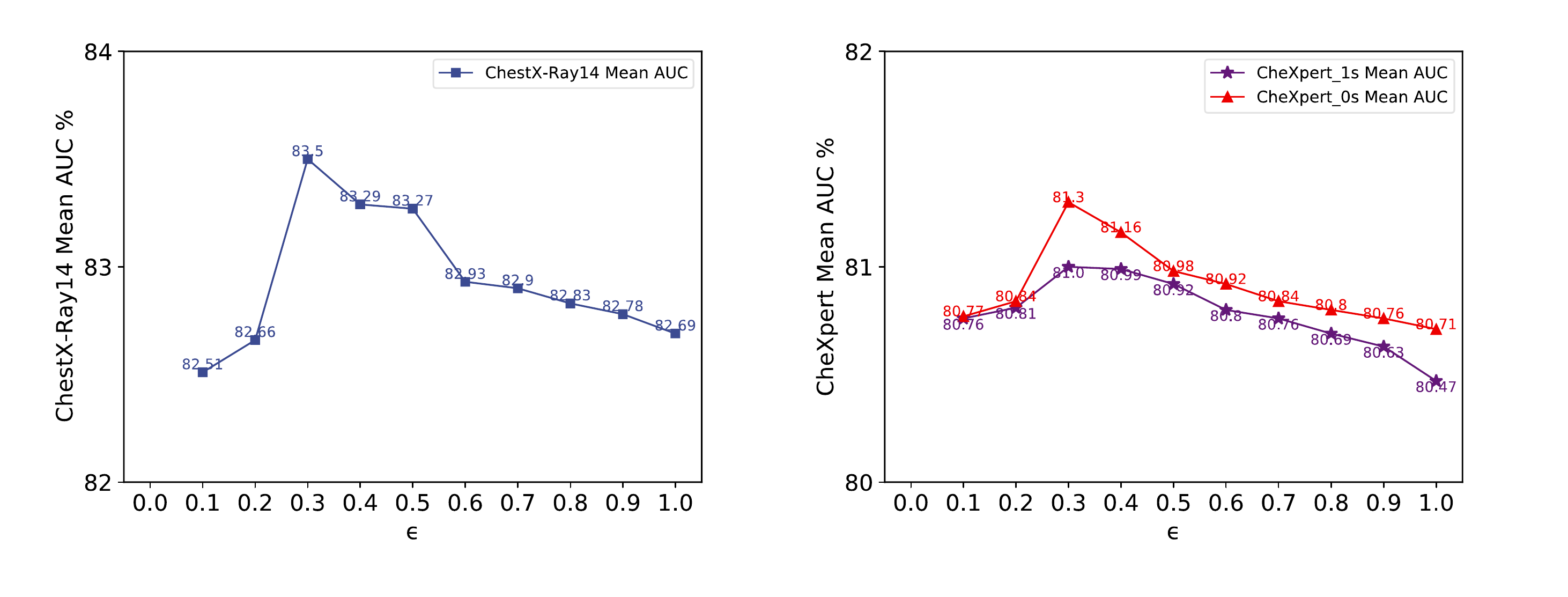}
  \caption{The change of Mean AUC using different values of $\epsilon$.}
  \label{fig:linjiejuzhen}
  \centering
  \includegraphics[width=\textwidth]{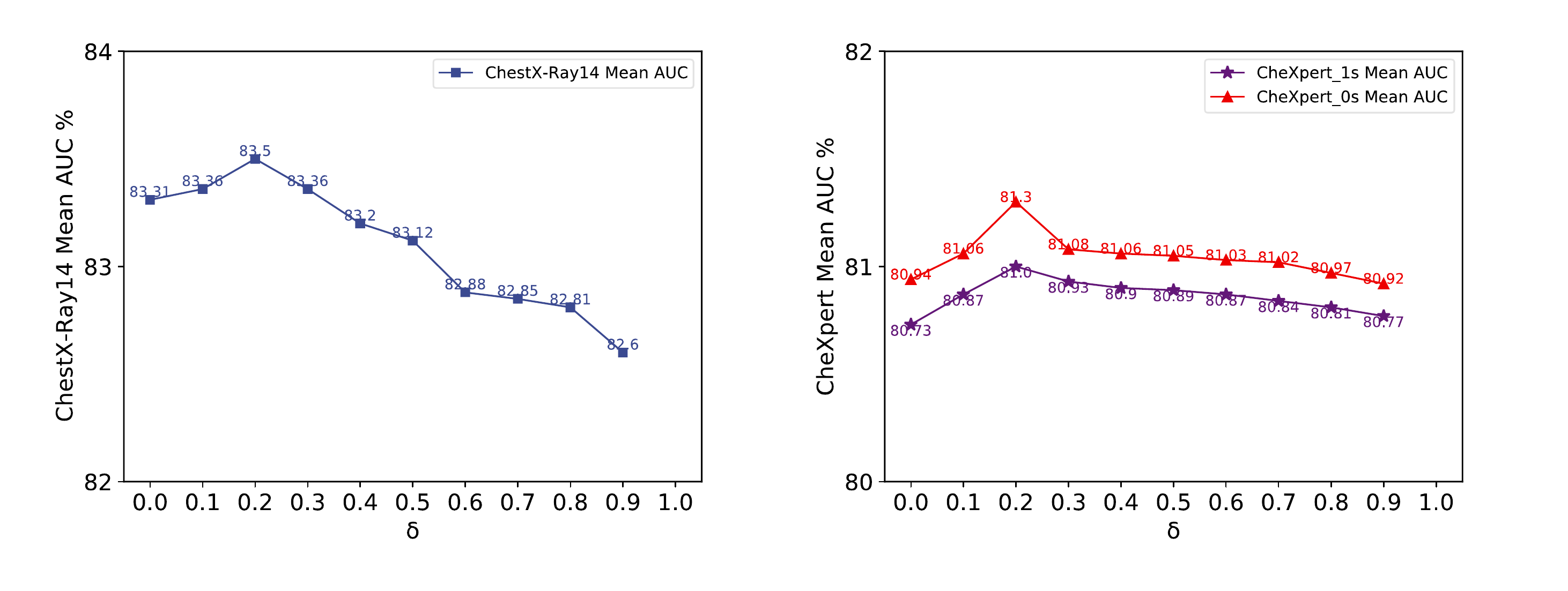} 
  \caption{The change of Mean AUC using different values of $\delta$. Note that, when $\delta$ = 1, the model does not converge.}
  \label{fig:jiaquanjuzhen}
  \centering
  \includegraphics[width=\textwidth]{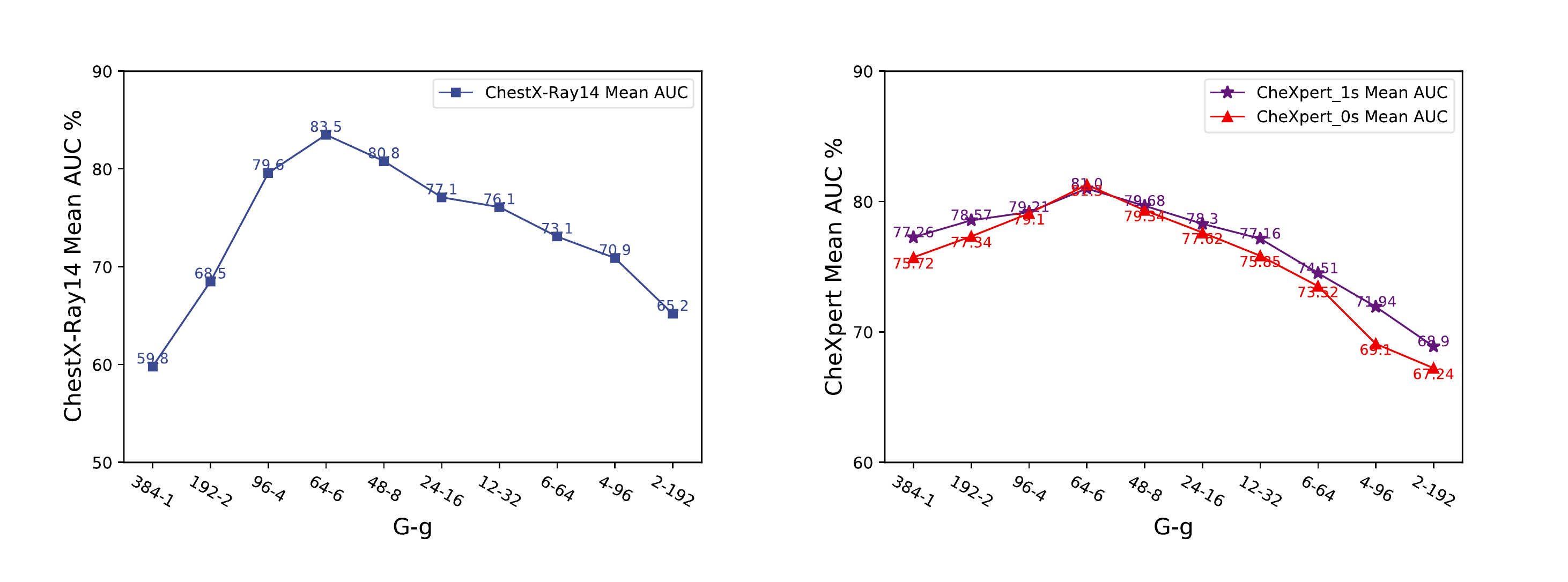} 
  \caption{The change of Mean AUC using different values of $G$-$g$.}
  \label{fig:G-g}
\end{figure}

In this section, we set $\epsilon$ by using
Eq.(\ref{equ:shizi4}) and $\delta$
Eq.(\ref{equ:shizi5}) with different values to construct a weighted pathology label co-occurrence correlation matrix to evaluate the performance of GCN.
The parameter $\epsilon$ filters noisy data (rare co-occurring pathological inter-label relationships) for a better model.
As shown in Fig.~\ref{fig:linjiejuzhen}, we incremented $\epsilon$ from 0.1 to 1 to observe the effect
and found that BB-GCN achieves the highest Mean AUC on ChestX-Ray14 and CheXpert when $\epsilon=0.3$. Note that when $\epsilon$=0, we do not filter any edges when constructing the pathology label correlation matrix. However, the model does not converge, so there is no result for $\epsilon$=0 in Fig.~\ref{fig:linjiejuzhen}.
This is because $\epsilon=0.3$ is a better balance parameter, which not only preserves the correlation between objects. Furthermore, it reduces the impact of the small probability of noisy data. In addition, the multi-label recognition accuracy of the two CXR datasets
we have used the following parameters in a set of
$\delta \in \left\{ 0 , 0.1 , 0.2 , \cdots , 0.9 , 1 \right\}$,
exploring the effect of different $\delta$ values in Eq.(\ref{equ:shizi5}) on the reweighting of the correlation matrix, as shown in Fig.~\ref{fig:jiaquanjuzhen}.
The figure shows the importance of balancing the weights between the node and neighboring nodes when updating the node features in GCN.
In our experiments, we choose the optimal value of $\delta$ by cross-validation.
We can see that when $\delta=$0.2, it achieves the best performance on both ChestX-Ray14 and CheXpert.
If $\delta$ is too tiny, the nodes (pathology labels) of the graph can not get enough information from the related nodes.
Moreover, if $\delta$ is too large, it will produce excessive smoothing.
The results show that $\delta=$0.2 can well balance the correlation of this inter-label relationship.
\begin{table*}[!htpb]
\centering
\caption{The different number of GCN layers in LCE.}
\label{tab:tab-gcn layer}
\resizebox{\textwidth}{!}{%
\begin{tabular}{cccccccccc}
\hline
         & \multicolumn{9}{c}{Dataset}                                                                                   \\ \cline{2-10} 
 \#layer   & \multicolumn{4}{c|}{CheXpert\_0s}             & \multicolumn{4}{c|}{CheXpert\_1s}             & ChestX-Ray14 \\ \cline{2-10} 
         & Mean AUC & OP & OR & \multicolumn{1}{c|}{OF1} & Mean AUC & OP & OR & \multicolumn{1}{c|}{OF1} & Mean AUC      \\ \hline
2-layers & 0.813       & 0.760 & 0.600 & \multicolumn{1}{c|}{0.665}  & 0.810       & 0.744 & 0.560 & \multicolumn{1}{c|}{0.649}  & 0.835          \\
3-layers & 0.811       & 0.748 & 0.572 & \multicolumn{1}{c|}{0.652}  & 0.808       & 0.736 & 0.537 & \multicolumn{1}{c|}{0.636}  & 0.833          \\
4-layers & 0.806       & 0.727 & 0.526 & \multicolumn{1}{c|}{0.621}  & 0.804       & 0.720 & 0.495 & \multicolumn{1}{c|}{0.619}  & 0.830          \\ \hline
\end{tabular}%
}
\end{table*}
\subsubsection{Different number of GCN layers in \textit{pathology Label Co-occurrence relationship Embedding} (LCE) module}
We show the performance results for different GCN layers of our model in Table~\ref{tab:tab-gcn layer}.
For the 3-layer model, the output dimensions of the sequential layers are 1024, 1024, and 768, respectively.
For the 4-layer model, the dimensions are 1024, 1024, 1024, and 768, respectively.
As shown in the table, the performance of multi-label recognition on both datasets decreases when the number of GCN layers increases.
The performance degradation is due to the accumulation of information transfer between nodes when more GCN layers are used, which leads to over-smoothing.

\subsubsection{Groups $G$ and Elements $g$ in GroupSum}
In this section, we evaluate the performance of the TBG bridging module in BB-GCN by using a different number of groups $G$ and the number of elements $g$ within a group.
With the GroupSum in the TBG bridging module, each $D_{3}$-dimensional vector will be converted into a $G$-dimensional vector.
We have a set of $G$-$g \in \{(1,384), (2,192), \cdots \}$
to generate a low-dimensional bridging vector.
As shown in Fig.~\ref{fig:G-g}, BB-GCN obtains better performance on ChestX-Ray14 when $G$=64 and $g$=6 are chosen, while the change of Mean AUC is very slight on CheXpert.
We believe the original semantic information between the pathology labels can be better expressed by $G$=64 and $g$=6.
Otherwise, other values of $G$-$g$ bring similar results, which do not affect the model too much.
\begin{figure}[!htp]
  \centering
  \includegraphics[height=0.42\textwidth, width=\textwidth]{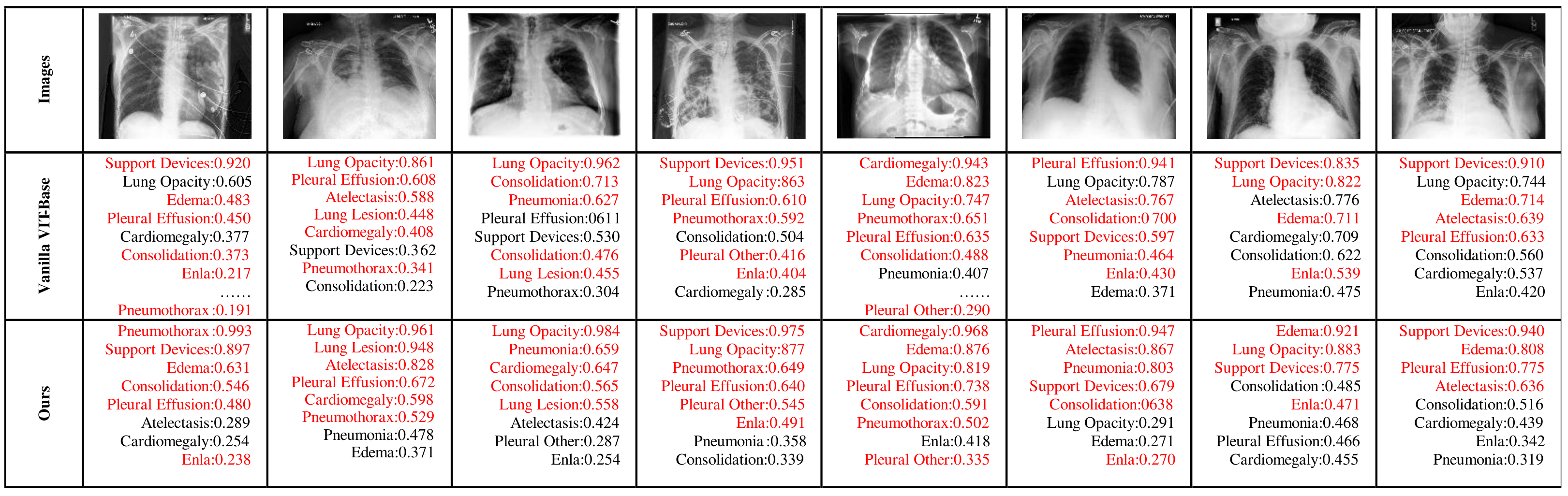} 
  \caption{Examples of recognition results. Present the top-8 predicted pathology labels and the corresponding probability scores. The ground truth labels are highlighted in red.}
  \label{fig:scores1}
\end{figure}
\subsection{Visualization of lesion areas for qualitative assessment}
Fig.~\ref{fig:scores1} illustrates a visual representation of multi-label CXR recognition.
The top-8 predicted scores for each test subject are given, sorted top-down by the magnitude of the predicted score values.
Compared with the vanilla ViT-Base model, the proposed BB-GCN enhances the performance of multi-label CXR recognition.
Our BB-GCN can effectively improve associated pathology confidence scores and suppress non-associated pathology scores with fully considered and modeled global label relationships.
For example, in column 2, the BB-GCN fully considers the pathological relationship between effusion and Atelectasis.
In the presence of effusion, the corresponding confidence score for Atelectasis was (0.588 $\rightarrow$ 0.828); compared to vanilla ViT-Base performance, the confidence score improved by about 0.240.
For the weakly correlated labels, the support device ranked 3rd in column 3 regarding the vanilla ViT-Base score.
While BB-GCN fully considers the global inter-label relationships, its confidence score does not reach the top 8.
To some extent, this demonstrates the ability of our model to suppress the confidence scores of non-relevant labels.

\section{Conclusion}
Assisting multi-label CXR recognition by capturing the complex relationships between pathology labels is essential for clinical diagnosis.
In order to capture the global co-occurrence relationship between pathology labels and efficiently fuse the vectors of cross-modalities, this paper proposes a bimodal bridging graph convolutional network, BB-GCN.
Specifically, our model first extracts feature representations of images via the Backbone module.
Then the LCE module is used to global model the pathology label correlation information and learns the pathology label co-occurrence relationship embedding.
Finally, the TBG module effectively bridged the two modal vectors (image feature representation and pathology label co-occurrence relationship embedding).
Extensive experimental results on ChestX-Ray14 and CheXpert show that the proposed LCE module and TBG module can effectively contribute to each other, thus significantly improving the multi-label CXR recognition performance of the model and achieving satisfactory results.
In the future, we will introduce attention mechanisms into the modeling of label relationships to learn more precise label relations and help further improve multi-label CXR recognition performance.

\section*{Acknowledgments}

This work is supported by the National Nature Science Foundation of China (No. 61872225), Introduction and Cultivation Program for Young Creative Talents in Colleges and Universities of Shandong Province (No.2019-173), the Natural Science Foundation of Shandong Province (No.ZR2020KF013, No.ZR2020ZD44, No.ZR2019ZD04, No.ZR2020QF043) and the Special fund of Qilu Health and Health Leading Talents Training Project.
\bibliography{cas-refs}

\begin{thebibliography}{10}
\expandafter\ifx\csname url\endcsname\relax
  \def\url#1{\texttt{#1}}\fi
\expandafter\ifx\csname urlprefix\endcsname\relax\def\urlprefix{URL }\fi
\expandafter\ifx\csname href\endcsname\relax
  \def\href#1#2{#2} \def\path#1{#1}\fi

\bibitem{hansell2008fleischner}
D.~M. Hansell, A.~A. Bankier, H.~MacMahon, T.~C. McLoud, N.~L. Muller, J.~Remy,
  et~al., Fleischner society: glossary of terms for thoracic imaging, Radiology
  246~(3) (2008) 697.

\bibitem{yao2017learning}
L.~Yao, E.~Poblenz, D.~Dagunts, B.~Covington, D.~Bernard, K.~Lyman, Learning to
  diagnose from scratch by exploiting dependencies among labels, arXiv preprint
  arXiv:1710.10501 (2017).

\bibitem{wang2017chestx}
X.~Wang, Y.~Peng, L.~Lu, Z.~Lu, M.~Bagheri, R.~M. Summers, Chestx-ray8:
  Hospital-scale chest x-ray database and benchmarks on weakly-supervised
  classification and localization of common thorax diseases, in: Proceedings of
  the IEEE conference on computer vision and pattern recognition, 2017, pp.
  2097--2106.

\bibitem{galleguillos2008object}
C.~Galleguillos, A.~Rabinovich, S.~Belongie, Object categorization using
  co-occurrence, location and appearance, in: 2008 IEEE Conference on Computer
  Vision and Pattern Recognition, IEEE, 2008, pp. 1--8.

\bibitem{shin2016learning}
H.-C. Shin, K.~Roberts, L.~Lu, D.~Demner-Fushman, J.~Yao, R.~M. Summers,
  Learning to read chest x-rays: Recurrent neural cascade model for automated
  image annotation, in: Proceedings of the IEEE conference on computer vision
  and pattern recognition, 2016, pp. 2497--2506.

\bibitem{clare2001knowledge}
A.~Clare, R.~D. King, Knowledge discovery in multi-label phenotype data, in:
  European conference on principles of data mining and knowledge discovery,
  Springer, 2001, pp. 42--53.

\bibitem{golfarelli2009survey}
M.~Golfarelli, S.~Rizzi, A survey on temporal data warehousing, International
  Journal of Data Warehousing and Mining (IJDWM) 5~(1) (2009) 1--17.

\bibitem{camnet}
M.~Li, M.~Wei, X.~He, F.~Shen, Enhancing pary features via contrastive
  attention module for vehicle re-identification, in: Conference on
  International Conference on Image Processing, IEEE, 2022.

\bibitem{zhang2013review}
M.-L. Zhang, Z.-H. Zhou, A review on multi-label learning algorithms, IEEE
  transactions on knowledge and data engineering 26~(8) (2013) 1819--1837.

\bibitem{wang2016cnn}
J.~Wang, Y.~Yang, J.~Mao, Z.~Huang, C.~Huang, W.~Xu, Cnn-rnn: A unified
  framework for multi-label image classification, in: Proceedings of the IEEE
  conference on computer vision and pattern recognition, 2016, pp. 2285--2294.

\bibitem{zhu2017learning}
F.~Zhu, H.~Li, W.~Ouyang, N.~Yu, X.~Wang, Learning spatial regularization with
  image-level supervisions for multi-label image classification, in:
  Proceedings of the IEEE conference on computer vision and pattern
  recognition, 2017, pp. 5513--5522.

\bibitem{wang2017multi}
Z.~Wang, T.~Chen, G.~Li, R.~Xu, L.~Lin, Multi-label image recognition by
  recurrently discovering attentional regions, in: Proceedings of the IEEE
  international conference on computer vision, 2017, pp. 464--472.

\bibitem{chen2019multi}
Z.-M. Chen, X.-S. Wei, P.~Wang, Y.~Guo, Multi-label image recognition with
  graph convolutional networks, in: Proceedings of the IEEE/CVF conference on
  computer vision and pattern recognition, 2019, pp. 5177--5186.

\bibitem{chen2020label}
B.~Chen, J.~Li, G.~Lu, H.~Yu, D.~Zhang, Label co-occurrence learning with graph
  convolutional networks for multi-label chest x-ray image classification, IEEE
  journal of biomedical and health informatics 24~(8) (2020) 2292--2302.

\bibitem{li2019learning}
Q.~Li, X.~Peng, Y.~Qiao, Q.~Peng, Learning category correlations for
  multi-label image recognition with graph networks, arXiv preprint
  arXiv:1909.13005 (2019).

\bibitem{dosoViTskiy2020image}
A.~Dosovitskiy, L.~Beyer, A.~Kolesnikov, D.~Weissenborn, X.~Zhai,
  T.~Unterthiner, M.~Dehghani, M.~Minderer, G.~Heigold, S.~Gelly, et~al., An
  image is worth 16x16 words: Transformers for image recognition at scale,
  arXiv preprint arXiv:2010.11929 (2020).

\bibitem{kipf2016semi}
T.~N. Kipf, M.~Welling, Semi-supervised classification with graph convolutional
  networks, arXiv preprint arXiv:1609.02907 (2016).

\bibitem{irvin2019chexpert}
J.~Irvin, P.~Rajpurkar, M.~Ko, Y.~Yu, S.~Ciurea-Ilcus, C.~Chute, H.~Marklund,
  B.~Haghgoo, R.~Ball, K.~Shpanskaya, et~al., Chexpert: A large chest
  radiograph dataset with uncertainty labels and expert comparison, in:
  Proceedings of the AAAI conference on artificial intelligence, Vol.~33, 2019,
  pp. 590--597.

\bibitem{krizhevsky2012imagenet}
A.~Krizhevsky, I.~Sutskever, G.~E. Hinton, Imagenet classification with deep
  convolutional neural networks, Advances in neural information processing
  systems 25 (2012).

\bibitem{simonyan2014very}
K.~Simonyan, A.~Zisserman, Very deep convolutional networks for large-scale
  image recognition, arXiv preprint arXiv:1409.1556 (2014).

\bibitem{szegedy2015going}
C.~Szegedy, W.~Liu, Y.~Jia, P.~Sermanet, S.~Reed, D.~Anguelov, D.~Erhan,
  V.~Vanhoucke, A.~Rabinovich, Going deeper with convolutions, in: Proceedings
  of the IEEE conference on computer vision and pattern recognition, 2015, pp.
  1--9.

\bibitem{he2016deep}
K.~He, X.~Zhang, S.~Ren, J.~Sun, Deep residual learning for image recognition,
  in: Proceedings of the IEEE conference on computer vision and pattern
  recognition, 2016, pp. 770--778.

\bibitem{ibrahim2021deep}
D.~M. Ibrahim, N.~M. Elshennawy, A.~M. Sarhan, Deep-chest: Multi-classification
  deep learning model for diagnosing covid-19, pneumonia, and lung cancer chest
  diseases, Computers in biology and medicine 132 (2021) 104348.

\bibitem{mahmud2020covxnet}
T.~Mahmud, M.~A. Rahman, S.~A. Fattah, Covxnet: A multi-dilation convolutional
  neural network for automatic covid-19 and other pneumonia detection from
  chest x-ray images with transferable multi-receptive feature optimization,
  Computers in biology and medicine 122 (2020) 103869.

\bibitem{ozturk2020automated}
T.~Ozturk, M.~Talo, E.~A. Yildirim, U.~B. Baloglu, O.~Yildirim, U.~R. Acharya,
  Automated detection of covid-19 cases using deep neural networks with x-ray
  images, Computers in biology and medicine 121 (2020) 103792.

\bibitem{rajpurkar2017chexnet}
P.~Rajpurkar, J.~Irvin, K.~Zhu, B.~Yang, H.~Mehta, T.~Duan, D.~Ding, A.~Bagul,
  C.~Langlotz, K.~Shpanskaya, et~al., Chexnet: Radiologist-level pneumonia
  detection on chest x-rays with deep learning, arXiv preprint arXiv:1711.05225
  (2017).

\bibitem{huang2017densely}
G.~Huang, Z.~Liu, L.~Van Der~Maaten, K.~Q. Weinberger, Densely connected
  convolutional networks, in: Proceedings of the IEEE conference on computer
  vision and pattern recognition, 2017, pp. 4700--4708.

\bibitem{shen2018dynamic}
Y.~Shen, M.~Gao, Dynamic routing on deep neural network for thoracic disease
  classification and sensitive area localization, in: International Workshop on
  Machine Learning in Medical Imaging, Springer, 2018, pp. 389--397.

\bibitem{sak2014long}
H.~Sak, A.~Senior, F.~Beaufays, Long short-term memory based recurrent neural
  network architectures for large vocabulary speech recognition, arXiv preprint
  arXiv:1402.1128 (2014).

\bibitem{ypsilantis2017learning}
P.-P. Ypsilantis, G.~Montana, Learning what to look in chest x-rays with a
  recurrent visual attention model, arXiv preprint arXiv:1701.06452 (2017).

\bibitem{tang2018attention}
Y.~Tang, X.~Wang, A.~P. Harrison, L.~Lu, J.~Xiao, R.~M. Summers,
  Attention-guided curriculum learning for weakly supervised classification and
  localization of thoracic diseases on chest radiographs, in: International
  Workshop on Machine Learning in Medical Imaging, Springer, 2018, pp.
  249--258.

\bibitem{guan2018diagnose}
Q.~Guan, Y.~Huang, Z.~Zhong, Z.~Zheng, L.~Zheng, Y.~Yang, Diagnose like a
  radiologist: Attention guided convolutional neural network for thorax disease
  classification, arXiv preprint arXiv:1801.09927 (2018).

\bibitem{lee2018multi}
C.-W. Lee, W.~Fang, C.-K. Yeh, Y.-C.~F. Wang, Multi-label zero-shot learning
  with structured knowledge graphs, in: Proceedings of the IEEE conference on
  computer vision and pattern recognition, 2018, pp. 1576--1585.

\bibitem{jia2023end}
H.~Jia, Z.~Xiao, P.~Ji, End-to-end fatigue driving eeg signal detection model
  based on improved temporal-graph convolution network, Computers in Biology
  and Medicine 152 (2023) 106431.

\bibitem{hsgm}
F.~Shen, X.~Peng, L.~Wang, X.~Zhang, M.~Shu, Y.~Wang, Hsgm: A hierarchical
  similarity graph module for object re-identification, in: 2022 IEEE
  International Conference on Multimedia and Expo (ICME), IEEE, 2022, pp. 1--6.

\bibitem{hpgn}
F.~Shen, J.~Zhu, X.~Zhu, Y.~Xie, J.~Huang, Exploring spatial significance via
  hybrid pyramidal graph network for vehicle re-identification, IEEE
  Transactions on Intelligent Transportation Systems (2021).

\bibitem{sun2022multi}
K.~Sun, M.~He, Y.~Xu, Q.~Wu, Z.~He, W.~Li, H.~Liu, X.~Pi, Multi-label
  classification of fundus images with graph convolutional network and
  lightgbm, Computers in Biology and Medicine 149 (2022) 105909.

\bibitem{marino2016more}
K.~Marino, R.~Salakhutdinov, A.~Gupta, The more you know: Using knowledge
  graphs for image classification, arXiv preprint arXiv:1612.04844 (2016).

\bibitem{wang2018zero}
X.~Wang, Y.~Ye, A.~Gupta, Zero-shot recognition via semantic embeddings and
  knowledge graphs, in: Proceedings of the IEEE conference on computer vision
  and pattern recognition, 2018, pp. 6857--6866.

\bibitem{yu2018modeling}
J.~Yu, Y.~Lu, Z.~Qin, W.~Zhang, Y.~Liu, J.~Tan, L.~Guo, Modeling text with
  graph convolutional network for cross-modal information retrieval, in:
  Pacific rim conference on multimedia, Springer, 2018, pp. 223--234.

\bibitem{fukui2016multimodal}
A.~Fukui, D.~H. Park, D.~Yang, A.~Rohrbach, T.~Darrell, M.~Rohrbach, Multimodal
  compact bilinear pooling for visual question answering and visual grounding,
  arXiv preprint arXiv:1606.01847 (2016).

\bibitem{xu2021dual}
R.~Xu, F.~Shen, H.~Wu, J.~Zhu, H.~Zeng, Dual modal meta metric learning for
  attribute-image person re-identification, in: 2021 IEEE International
  Conference on Networking, Sensing and Control (ICNSC), Vol.~1, IEEE, 2021,
  pp. 1--6.

\bibitem{kim2016hadamard}
J.-H. Kim, K.-W. On, W.~Lim, J.~Kim, J.-W. Ha, B.-T. Zhang, Hadamard product
  for low-rank bilinear pooling, arXiv preprint arXiv:1610.04325 (2016).

\bibitem{yu2018beyond}
Z.~Yu, J.~Yu, C.~Xiang, J.~Fan, D.~Tao, Beyond bilinear: Generalized multimodal
  factorized high-order pooling for visual question answering, IEEE
  transactions on neural networks and learning systems 29~(12) (2018)
  5947--5959.

\bibitem{pennington2014glove}
J.~Pennington, R.~Socher, C.~D. Manning, Glove: Global vectors for word
  representation, in: Proceedings of the 2014 conference on empirical methods
  in natural language processing (EMNLP), 2014, pp. 1532--1543.

\bibitem{maas2013rectifier}
A.~L. Maas, A.~Y. Hannun, A.~Y. Ng, et~al., Rectifier nonlinearities improve
  neural network acoustic models, in: Proc. icml, Vol.~30, Atlanta, Georgia,
  USA, 2013, p.~3.

\bibitem{lin2015bilinear}
T.-Y. Lin, A.~RoyChowdhury, S.~Maji, Bilinear cnn models for fine-grained
  visual recognition, in: Proceedings of the IEEE international conference on
  computer vision, 2015, pp. 1449--1457.

\bibitem{deng2009imagenet}
J.~Deng, W.~Dong, R.~Socher, L.-J. Li, K.~Li, L.~Fei-Fei, Imagenet: A
  large-scale hierarchical image database, in: 2009 IEEE conference on computer
  vision and pattern recognition, Ieee, 2009, pp. 248--255.

\bibitem{guendel2018learning}
S.~Guendel, S.~Grbic, B.~Georgescu, S.~Liu, A.~Maier, D.~Comaniciu, Learning to
  recognize abnormalities in chest x-rays with location-aware dense networks,
  in: Iberoamerican Congress on Pattern Recognition, Springer, 2018, pp.
  757--765.

\bibitem{chen2019dualchexnet}
B.~Chen, J.~Li, X.~Guo, G.~Lu, Dualchexnet: dual asymmetric feature learning
  for thoracic disease classification in chest x-rays, Biomedical Signal
  Processing and Control 53 (2019) 101554.

\end{thebibliography}

\end{sloppypar}
\end{document}